# General-Kindred Physics-Informed Neural Network to the Solutions of Singularly Perturbed Differential Equations

Sen Wang,1 Peizhi Zhao,1 Qinglong Ma,1 and Tao Song∗1

*College of Computer Science and Technology, China University of Petroleum (East China), Qingdao 266580, P. R. China*

(*Electronic mail: tsong@upc.edu.cn)

(Dated: 27 August 2024)

Physics-Informed Neural Networks (PINNs) have become a promising research direction in the field of solving Partial Differential Equations (PDEs). Dealing with singular perturbation problems continues to be a difficult challenge in the field of PINN. The solution of singular perturbation problems often exhibits sharp boundary layers and steep gradients, and traditional PINN cannot achieve approximation of boundary layers. In this manuscript, we propose the General-Kindred Physics-Informed Neural Network (GKPINN) for solving Singular Perturbation Differential Equations (SPDEs). This approach utilizes asymptotic analysis to acquire prior knowledge of the boundary layer from the equation and establishes a novel network to assist PINN in approximating the boundary layer. It is compared with traditional PINN by solving examples of one-dimensional, two-dimensional, and time-varying SPDE equations. The research findings underscore the exceptional performance of our novel approach, GKPINN, which delivers a remarkable enhancement in reducing the $L_2$ error by two to four orders of magnitude compared to the established PINN methodology. This significant improvement is accompanied by a substantial acceleration in convergence rates, without compromising the high precision that is critical for our applications. Furthermore, GKPINN still performs well in extreme cases with perturbation parameters of $1 \times 10^{-38}$, demonstrating its excellent generalization ability.

## I. INTRODUCTION

Singular perturbation differential equations are a specialized type of differential equation with wide-ranging applications in fluid mechanics. The coefficient of the highest derivative in SPDEs is typically denoted by the positive parameter $\varepsilon$, often referred to as the perturbation parameter. As $\varepsilon$ approaches 0, the solution of these equations undergoes rapid changes at the boundary layer, also known as the thin region. Hence, traditional numerical methods using uniform grids face challenges in approximating solutions for solving SPDEs.[1] This poses significant obstacles to the resolution of SPDE problems.

The emergence and advancement of deep learning have sparked a growing interest in the utilization of artificial neural networks (ANN) for solving partial differential equations.[2] Deep learning methods, such as physics-informed neural networks (PINNs), have emerged as versatile approaches for solving partial differential equations.[3–5] PINN utilizes neural networks to approximate the solution of the desired PDE and incorporate residual terms into the loss function to train and optimize the network by minimizing the residuals of the PDE.[6,7] In recent years, there have been numerous studies on PINN, primarily focused on designing more effective neural network architectures and advanced training algorithms to enhance the performance of PINN. For instance, adaptive resampling techniques such as adaptively re-sampling collocation points can be employed to balance the different loss terms in the loss function and enhance training accuracy. Examples of these techniques include importance sampling, evolutionary sampling, residual-based adaptive sampling, etc.[8–10] Similarly, alternative methods such as loss re-weighting schemes adjust the weight of loss during training to achieve similar objectives. These methods include non-adaptive weighting, CasualPINN, RBA-PINN, PINN-NTK, SA-PINN, and the learning rate annealing algorithm.[11–16] Researchers have also endeavored to develop innovative neural network architectures to enhance the performance of PINN. These efforts include the incorporation of positional embeddings such as Fourier feature embeddings,[17] as well as the exploration of novel architectures.[18–21] Additionally, other studies have approached this challenge from alternative perspectives, such as the integration of finite element analysis with PINN.[22,23] Some methods employ the concept of constraint optimization to reframe the loss function as a problem of maximizing minimum, which also demonstrates strong predictive potential.[24] However, PINN still encounters challenges when solving singular perturbation differential equations. Traditional PINN struggles to accurately approximate and accommodate the sharp changes in the boundary layer when the perturbation parameter is less than $1 \times 10^{-3}$. In this study, we propose a novel approach called the general-kindred physics-informed neural network(GKPINN), which has been specifically designed to address these issues in the boundary layer. We integrate prior knowledge of progressive analysis into neural networks and theoretically decompose SPDEs into smooth and layer components to achieve effective problem-solving. Our contributions are threefold:

## II. RELATED WORK

### A. Physics-informed neural networks

Traditional machine learning models primarily rely on large volumes of data for training. In many scenarios, data is either limited or the cost of acquisition is high, despite the significant breakthroughs that data-driven machine learning has achieved in various fields. As a result, researchers are exploring the integration of physics knowledge with machine learning to address these challenges. The integration of data and physical models in machine learning, based on physical information, has emerged as a key area of research. This line of inquiry extends to PINN and other PDE solvers.[3,25–28] Among these, PINN has had the most significant impact.

PINN takes into account physical laws during the training process and integrates physical equations into the loss function, which enhances their generalization ability. This feature allows PINN to be trained in situations where data is limited or even unavailable, making them suitable for specific application scenarios. However, it is undeniable that PINN also face numerous challenges: PINN lacks accuracy and often presents difficulties in training for solving multi-scale problems. In recent years, researchers have conducted detailed studies on the origins of these issues and potential solutions, demonstrating greater promise for PINN.[14,16,29]

Moving forward, we will provide a brief overview of the PINN in the context of solving PDEs, and utilize the following partial differential equations as examples:

$$u_t - \varepsilon u_{xx} - u_x + 5u = 0, \; x,t \in (0, 1) \quad (1)$$

The initial and boundary conditions are as follows:

$$u(x, 0) = sin(2\pi x) \quad (2)$$

$$u(0,t) = 0, \; u(1,t) = 1 \quad (3)$$

Represent the unknown solution $u(x,t)$ using a deep neural network $u_\theta(x,t)$, where $\theta$ represents all adjustable parameter weights $W$ and biases $b$ in the neural network. Therefore, PDE residuals can be defined as:

$$R_\theta(x,t) = \frac{\partial u_\theta}{\partial t}(x_r,t_r) + N[u_\theta](x_r,t_r) \quad (4)$$

Subsequently, the physical model should be trained by minimizing the loss function:

$$L(\theta) = L_{ic}(\theta) + L_{bc}(\theta) + L_r(\theta) \quad (5)$$

Where:

$$L_{ic}(\theta) = \frac{1}{N_{ic}} \sum_{i=1}^{N_{ic}} |u_\theta(x_{ic}, 0) - sin(2\pi x_{ic})|^2 \quad (6)$$

$$L_{bc}(\theta) = \frac{1}{N_{bc}} \sum_{i=1}^{N_{bc}} |u_\theta(0,t_{bci})|^2 + |u_\theta(1,t_{bci}) - 1|^2 \quad (7)$$

$$L_r(\theta) = \frac{1}{N_r} \sum_{i=1}^{N_r} |R_\theta(x_{ri},t_{ri})|^2 \quad (8)$$

Here $N_{ic}$, $N_{bc}$, and $N_r$ represent the initial training data, boundary training data, and the number of internal collocation points, respectively. Here $(x_{ic}, 0)$ denotes the initial condition point used as input at $t = 0$, and $(0,t_{bci})$ represents the boundary condition point when $x$ is located on both sides of the boundary. Additionally, $(x_{ir},t_{ri})$ signifies the collocation point passed to the residual $R_\theta(x,t)$. The required gradients for input variables and parameter $a$ in the neural network can be effectively calculated through automatic differentiation.

B. Residual-based attention

During the process of training neural networks, a persistent issue has emerged: when calculating the loss function, conventional neural networks tend to only consider the overall mean of residuals and overlook the point-by-point errors of residuals. This limitation often results in an incomplete capture of spatial or temporal features. The optimal approach to addressing this issue is to select a set of global or local weight multipliers. Among these, techniques such as adaptive weighting, causal training, and residual-based attention (RBA) weighting have demonstrated outstanding performance in PINN and other supervised learning tasks. RBA weights are based on residual exponential weighted moving averages, which effectively eliminate poor local minima or saddle points while capturing the spatial and temporal characteristics of specific problems. In comparison to traditional PINN, this method significantly reduces relative errors and accelerates convergence speed.

The update rule for RBA for any training point $i$ on iteration $k$ is given by:

$$\alpha_i^{k+1} \leftarrow (1 - \eta_*) \alpha_i^k + \eta_* \frac{|e|}{||e||_\infty}, \; i \in \{0, 1,..., N\} \quad (9)$$

where $N$ is the number of training points, $e_i$ is the residual of the respective loss term for point $i$ and $\eta_*$ is a learning rate.

III. PROBLEM SETTINGS AND METHOD

A. Singularly perturbed differential equations

Singular perturbation differential equations (SPDEs) involve a small positive parameter, typically denoted as $\varepsilon$, which commonly appears in front of the highest-order derivative term. As the parameter $\varepsilon$ approaches 0, significant changes will occur in certain regions of the solution, and the derivative of the solution may tend towards infinity. These specific regions are known as boundary layers. The location of the boundary layer is determined by the specific form of the equation. Here is a basic one-dimensional convection-diffusion equation as an illustration:

$$-\varepsilon u_{xx} + u_x = \varepsilon \pi^2 sin(\pi x) + \pi cos(\pi x), \; x \in (0, 1) \quad (10)$$
$$u(0) = 0, \; u(1) = 1$$

Where $\varepsilon$ is a very small positive parameter. Due to the influence of $\varepsilon$, the solution of this equation exhibits singular behavior at $x = 1$, as illustrated in Figure 1.

However, as $\varepsilon$ continues to shrink, the changes in the solution at the boundary layer will become more significant. This has presented great challenges to the solution work of PINN, as illustrated in Figure 2.

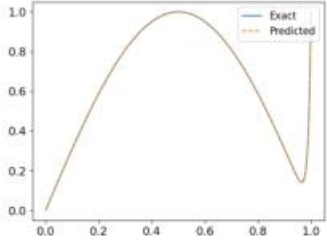

FIG. 1. Ground truth and predictions for PINN. Here, we set $\varepsilon = 0.01$ in Eq. (10) and substitute 1000 collocation points obtained from Latin hypercube sampling into the training. It can be observed that the PINN has effectively achieved fitting.

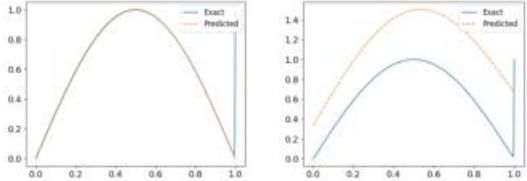

FIG. 2. Ground truth and predictions for PINN when $\varepsilon = 0.001$ in Eq. (10). The left and right images display the training results of PINN under supervised and unsupervised learning respectively. Even under data-driven conditions, it is evident that the PINN is unable to learn steep gradients at the boundary layer when the parameters are set small enough. It is precisely because of this that when PINN is under unsupervised learning conditions, it cannot effectively fit the entire solution.

### B. Asymptotic analysis

For SPDEs, their solutions are typically decomposed into two components. One part is utilized to represent the behavior of the solution in the boundary region, referred to as the layer part. The other part pertains to the behavior of solutions in regions other than the boundaries, known as the smoothing part. It is well-known that as the perturbation parameter $\varepsilon$ approaches 0, there will be a rapid change in the boundary layer region. Numerous studies utilize the concept of asymptotic analysis to solve SPDEs.[1,30] For rapid changes within the boundary layer, we can approximate them using an exponential function. The solution within the boundary layer can be represented in the form of an exponential layer, which provides conceptual support for our innovative architecture. Our framework is constructed based on singular perturbation theory and matched asymptotic expansion. Singular perturbation theory requires the matching of solutions within the boundary layer (exponential layer) with those outside the boundary layer (asymptotic layer). In the following section, we will demonstrate the phenomena of boundary and inner layers in various types of SPDEs through specific examples.

For one-dimensional equations:

$$-\varepsilon u_{xx} + b(x)u_x + c(x)u = f(x), x \in (0, 1) \quad (11)$$
$$u(0) = 0, u(1) = 1$$

In the domain of $x$, where $b(x) > 0$, an exponential boundary layer typically emerges at $x = 1$. Similarly, if $b(x) < 0$, the boundary layer will occur at $x = 0$. We use $u_{ax}$ to represent the m-order asymptotic expansion of $u(x)$, where there exists a constant $C$ and a sufficiently small $\varepsilon$. For any $x \in [0, 1]$:

$$|u(x) - u_{ax}| \leq C\varepsilon^{m+1} \quad (12)$$

According to the theory of singular perturbation, the solution can be divided into two parts: the smooth part is denoted as $u_0$, and the layer part is denoted as $v_0$. This implies that:

$$|u(x) - (u_0 + v_0)| \leq C\varepsilon, m = 0 \quad (13)$$

Where, through the matching of asymptotic expansion, we can see in the current example:

$$v_0(x) = (u(1) - u_0(1))e^{-b(1)\frac{1-x}{\varepsilon}} \quad (14)$$

For the time-dependent equation, $Q = (0, 1) \times (0, T]$, we have:

- $u_t - \varepsilon u_{xx} + b(x,t)u_x + c(x,t)u = f(x,t), (x,t) \in Q$
- $u(x, 0) = g(x)$ \quad (15)
- $u(0,t) = q_0(t), u(1,t) = q_1(t)$

When dealing with such equations, the approach is very similar to Eq. (11). In cases where $b > 0$, the solution $u$ of the equation generally demonstrates smooth behavior in the region $Q$. But as $x$ approaches the boundary $x = 1$, $u$ will display a sharp boundary phenomenon.

We can expand the traditional PINN to the structure depicted in Figure 3 with the support of the current approach. GKPINN incorporates prior knowledge of layer components obtained through asymptotic analysis, dividing the architecture into two blocks. The first part utilizes fully connected layers to fit the smooth components in SPDEs. The second part involves multiplying the second fully connected layer with the exponential layer, which is determined by the position and type of the boundary layer, to capture specific information related to the layer. Simply put, if we consider the deep neural network $u_\theta$ in PINN as $u(x)$, then our task is to transform $u_\theta$ into the required form $u_{ax}$ to achieve learning of the boundary layer. For spatial dimensions greater than or equal to 2, such as in the case of two-dimensional PDE equations, there may be instances in which both the $x$ and $y$ dimensions exhibit boundary layers.

Among them, we define the mathematical set $\Omega = (0, 1)^2$, we have:

$$-\varepsilon\Delta u + b(x, y)\nabla u + c(x, y)u = f(x, y), in\ \Omega \quad (16)$$
$$u(x, y) = 0, on\ \partial\Omega$$

In this equation, the values of $b_1$ and $b_2$ in $b = b(x, y) = (b_1, b_2)$ should be considered separately. If $b = (b_1, b_2) > 0$ in Eq. (16), two exponential boundary layers will appear at x=1 and y=1 simultaneously. For the newly added boundary layer y=1, a new network must be established and paired with corresponding exponential layers for further fitting. Therefore, the model's architecture can be described as:

$$GKPINN = u_0 + \sum_{i=1}^{N} u_i * exp(-\alpha_i) \quad (17)$$

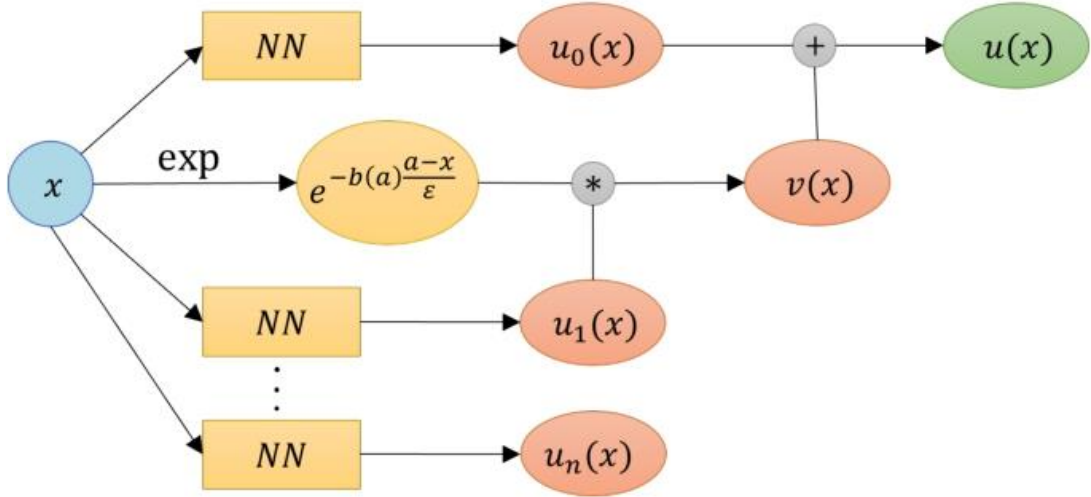

FIG. 3. GKPINN architecture for solving one-dimensional SPDEs. Where $x$ represents the input of the model, $NN$ denotes the fully connected layer, $a$ signifies the location of the boundary layer, $b(x)$ stands for the coefficient of $u_x$, and $exp$ indicates the exponential operation. For the boundary layer in the equation, we will introduce a new function $u_1(x)$ and multiply it with the previously obtained prior knowledge $e^{-b(a)\frac{a-x}{\varepsilon}}$, incorporating it into a function $v(x)$ that satisfies the boundary layer. If we extend our analysis to other SPDEs, each time a new boundary layer arises, a new network and exponential layer will be required to process it. Furthermore, the networks associated with different boundary layers will operate independently of one another.

Where $u_0$ represents the smooth component of the solution to the equation, which is represented using a fixed network. $N$ denotes the number of boundary layers, and $exp(-\alpha_i)$ refers to the exponential layers processed by different boundary layers, which vary depending on their position and type. Using Eq. (16) as an illustration:

$$\alpha_1 = b(1, y)(1 - x)/\varepsilon \qquad (18)$$
$$\alpha_2 = b(x, 1)(1 - y)/\varepsilon$$

Where $u_i$ represents a neural network combined with the exponential layer $exp(-\alpha_i)$.

It is important to note that when utilizing the concept of matching asymptotic expansion,[31] it may be necessary to ensure a smooth transition of solutions at the boundary layer by incorporating correction terms, particularly for problems with intricate boundaries. Especially when encountering situations similar to Eq. (16), the boundary layers $x = 1$ and $y = 1$ will overlap at (1, 1), resulting in corner layers. This may lead to certain instability. If another network is included for calibration, it not only increases the learning cost but also adds complexity to the network. We rely on neural networks to autonomously address this issue. It is evident that our architecture has modified the coefficient $u(1) - u_1(1)$ before the exponent to a neural network $u_1$ that is independent of $u_0$, upon observing the expansion of Eq. (14). This modification aims to take advantage of automatic differentiation's capability to gradually minimize the impact of correction terms during the network training process. The subsequent experimental section will illustrate the effectiveness of this architectural approach.

## IV. RESULTS AND DISCUSSION

This section will assess the performance of GKPINN in various scenarios of SPDEs, including ordinary differential equations and partial differential equations. The selected perturbation parameters $\varepsilon$ in our experiment are $1 \times 10^{-3}$ and $1 \times 10^{-38}$. The former will be compared and analyzed with the results obtained using traditional PINN, while the latter will be used to verify the generalization of GKPINN in handling SPDEs.

The network architecture comprises of two fully connected layers, each consisting of 100 neurons. The optimizer utilized in the network is Adam, with a learning rate set at 0.001. In the experiment of one-dimensional equations, the Sigmoid activation function is utilized. A total of 1000 points are sampled from Latin hypercubes, with an additional 50 randomly selected boundary points. In addition, 400 points were uniformly selected and substituted into the analytical solution to create a test set; In the experiment involving two-dimensional and time-dependent equations, we utilized the Tanh activation function and matched 10000 points obtained from Latin hypercube sampling. The initial and boundary points, each consisting of 100 randomly selected points, are used. The test set is obtained by employing high-precision finite difference methods. We initialize the RBA weights to 1 and update them with a learning rate of $\eta_* = 0.0001$, as described in Eq. (9).

For problems of varying dimensions, the randomness of point selection makes it challenging to make a reasonable and fair judgment on the training effect by solely considering training loss and $L_1$ error. Therefore, upon completion of the training, the effectiveness of the training is assessed using the $L_2$ norm of the test error:

$$L_2 = \|\hat{u} - u\|_2 = \sqrt{\frac{\sum_{i=1}^{N_{test}} |\hat{u}(x_i) - u(x_i)|^2}{\sum_{i=1}^{N_{test}} |\hat{u}(x_i)|^2}} \quad (19)$$

Where $N_{test}$ represents the number of points in the test set, and $\hat{u}$ denotes the approximate solution obtained through deep learning.

### A. Ordinary Differential Equations

For one-dimensional equations:

$$-\varepsilon u'' + b(x)u' + c(x)u = f(x), \quad x \in (0, 1) \quad (20)$$

From the concept of asymptotic analysis, we can acquire prior knowledge regarding the boundary layer:

$$b(x) > 0 \rightarrow x = 1 \quad (21)$$
$$b(x) < 0 \rightarrow x = 0$$

Therefore, we will classify and discuss based on the different positions of the boundary layer in one-dimensional equations. The initial stage of our exploration is framed by the following questions:

**Example1**

The first question selects Eq. (10), for which an analytical solution exists:

$$u(x) = \sin(\pi x) + \frac{e^{\frac{x}{\varepsilon}} - 1}{e^{\frac{1}{\varepsilon}} - 1} \quad (22)$$

This issue was previously mentioned in the Problem Settings section. Based on prior knowledge, it is understood that the boundary layer of the solution to this problem is located at $x = 1$. Therefore, an exponential layer $e^{(x-1)/\varepsilon}$ is utilized to match the fully connected layer $u_1$ to handle steep changes at the boundary layer. The experimental results are depicted in Figure 4.

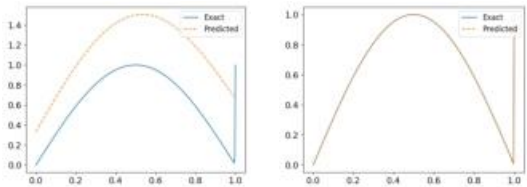

FIG. 4. Ground truth and predictions for PINN and GKPINN when $\varepsilon = 0.001$ in Eq. (10). It can be seen that there are discrepancies between the two models regarding this issue.

As observed, in contrast to the previous two learning methods which were ineffective with PINN, GKPINN successfully achieved effective fitting at the boundary layer.

**Example2**

Here we try another scenario:

$$-\varepsilon u_{xx} + u_x + (1 + \varepsilon)u = 0, \quad x \in (0, 1) \quad (23)$$
$$u(0) = 1 + e^{-\frac{1}{1+\varepsilon}}, \quad u(1) = 1 + e^{-1}$$

There exists an analytical solution to this problem:

$$u(x) = e^{-x} + e^{\frac{(1+\varepsilon)(x-1)}{\varepsilon}} \quad (24)$$

The boundary layer of the solution to this problem is also located at $x = 1$, therefore the same method as in Example 1 can be applied. The experimental results are depicted in Figure 5.

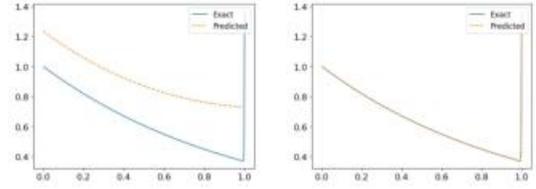

FIG. 5. Ground truth and predictions for PINN and GKPINN when $\varepsilon = 0.001$ in Eq. (24).

It can be seen that GKPINN effectively achieves fitting, however, due to PINN's inability to approximate the boundary layer, there are challenges in learning the entire solution.

**Example3**

Subsequently, we shall delve into the scenario in which the boundary layer is precisely located at the coordinate $x = 0$:

$$\varepsilon u_{xx} + (1 + \varepsilon)u_x + u = 0, \quad x \in (0, 1) \quad (25)$$
$$u(0) = 0, \quad u(1) = 1$$

There exists an analytical solution to this problem:

$$u(x) = \frac{e^{-x} - e^{-\frac{x}{\varepsilon}}}{e^{-x} - e^{-\frac{1}{\varepsilon}}} \quad (26)$$

Multiply the exponential layer $e^{-x/\varepsilon}$ by the output $u_1$ of the second network to create a network architecture capable of handling the boundary layer $x = 0$. As depicted in Figure 6, which outcomes are akin to Example 1,2, GKPINN yielded outstanding results; However, the training results were suboptimal due to PINN's incapacity to accurately model the curve at $x = 0$.

### B. Partial Differential Equations

Initially, we consider the two-dimensional equation:

$$-\varepsilon \Delta u + b(x, y)\nabla u + c(x, y)u = f(x, y) \quad (27)$$

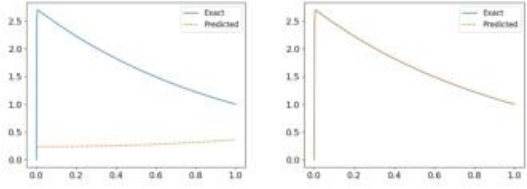

FIG. 6. Ground truth and predictions for PINN and GKPINN when $\varepsilon = 0.001$ in Eq. (25).

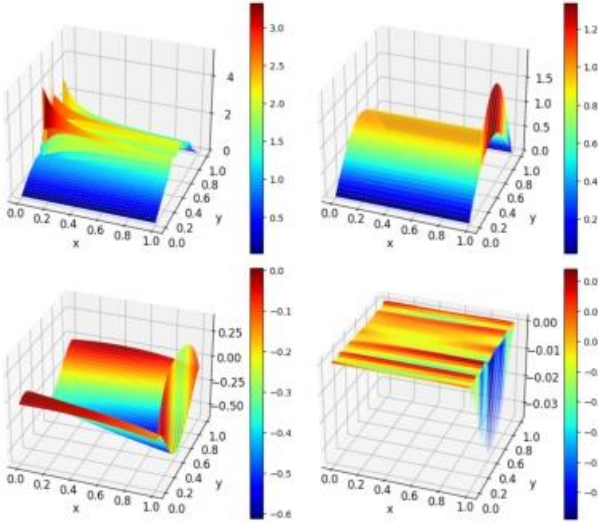

FIG. 7. When $\varepsilon = 0.001$ in Eq. (29), above is the prediction image of PINN(left) and GKPINN(right), and below is the error of PINN(left) and GKPINN(right) to the exact solution.

In this case, let $b(x, y) = b = (x_1, x_2)$. We can determine the location of the boundary layer with the concept of asymptotic analysis:

$$\begin{cases} b_i = 0 \rightarrow no\ layer \\ b_i > 0 \rightarrow right\ layer \\ b_i < 0 \rightarrow left\ layer \end{cases} \quad (28)$$

Example4

In two-dimensional equations, the boundary layer becomes more intricate. In this regard, we will explore various scenarios:

$$\begin{cases} -\varepsilon(u_{xx} + u_{yy}) + u_x = 0, (x, y) \in (0, 1)^2 \\ u(x, 0) = u(x, 1) = 0 \\ u(0, y) = sin(\pi y), u(1, y) = 2sin(\pi y) \end{cases} \quad (29)$$

For the given equation, let $b = (b_1, b_2) = (1, 0)$ indicate that the boundary layer is located at $x = 1$. Multiply $e^{(x-1)/\varepsilon}$ by $u_1$ output from the second network to process the layer part of SPDEs. Figure 7 illustrates the predicted images and errors of PINN and GKPINN. It is evident that as $x$ approaches 1, a steep slope transient occurs, and PINN fails to capture the boundary layer, leading to instability during training.

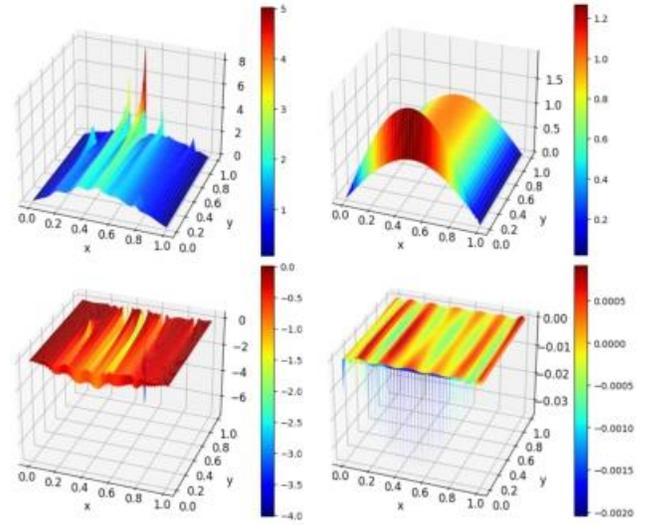

FIG. 8. When $\varepsilon = 0.001$ in Eq. (30), above is the prediction image of PINN(left) and GKPINN(right), and below is the error of PINN(left) and GKPINN(right) to the exact solution.

Example5

Subsequently, we shall contemplate an alternative boundary condition scenario:

$$\begin{cases} \varepsilon(u_{xx} + u_{yy}) + u_y = 0, (x, y) \in (0, 1)^2 \\ u(0, y) = u(1, y) = 0 \\ u(x, 0) = sin(\pi x), u(x, 1) = 2sin(\pi x) \end{cases} \quad (30)$$

Compared to Example 4, this example has simply adjusted the position of the boundary layer. In this equation, $b = (b_1, b_2) = (0, -1)$, therefore the boundary layer is located at $y = 0$. The same approach as Example 4 can be used to handle the layer part of the problem by multiplying $e^{-y/\varepsilon}$ with $u_1$ output by the second network. Figure 8 shows the results of PINN and GKPINN.

Example6

The following example is designed to test the scenario in which boundary layers appear in both dimensions:

$$\begin{cases} \varepsilon(u_{xx} + u_{yy}) + u_x + u_y = 0, (x, y) \in (0, 1)^2 \\ u(0, y) = u(1, y) = 0 \\ u(x, 0) = sin(\pi x), u(x, 1) = 2sin(\pi x) \end{cases} \quad (31)$$

For this particular problem, let $b = (b_1, b_2) = (-1, -1)$, indicating the existence of boundary layers at $x = 0$ and $y = 0$, respectively. We multiply $e^{-x/\varepsilon}$ by the second component of network $u_1$ to address the boundary layer at $x = 0$, and we multiply $e^{-y/\varepsilon}$ by $u_2$ to train the boundary layer that fits $y = 0$. The experimental results are depicted in Figure 9. In the presence of boundary layers on both sides, GKPINN is still able to achieve fitting, demonstrating the effectiveness of our method in solving two-dimensional equations.

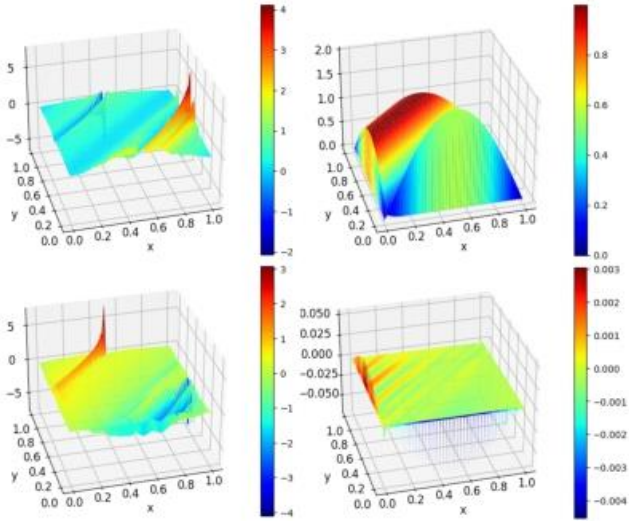

FIG. 9. When ε = 0.001 in Eq. (31), above is the prediction image of PINN(left) and GKPINN(right), and below is the error of PINN(left) and GKPINN(right) to the exact solution.

Example7

Finally, we will examine the differential equations in the time domain:

$$u_t - \varepsilon u_{xx} + b(x,t)u_x + c(x,t)u = f(x,t) \tag{32}$$

Our method of judgment for time-related equations closely resembles the process of solving one-dimensional equations:

$$\begin{aligned} b(x,t) > 0 &\to x = 1 \\ b(x,t) < 0 &\to x = 0 \end{aligned} \tag{33}$$

Proceeding further, experiments will be conducted on two distinct scenarios:

- $u_t - \varepsilon u_{xx} - u_x - u = 0, (x,t) \in (0, 1) \times (0, 1]$
- $u(x, 0) = cos(2\pi x)$
  $u(0,t) = 0, u(1,t) = 1$ (34)

For this equation, our objective is for the PINN to accurately capture the steep gradient at $x = 0$. This can be achieved by incorporating $e^{-x/\varepsilon}$ with $u_1$. The predicted images of PINN and GKPINN are shown in Figure 10. It can be seen that GKPINN effectively captures a steep gradient resembling a cross-section at $x = 0$, while the fitting effect of PINN in this equation is still not ideal.

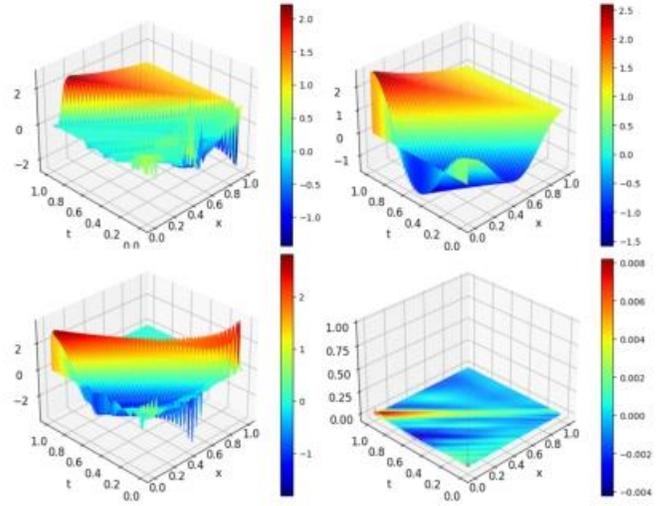

FIG. 10. When ε = 0.001 in Eq. (35), above is the prediction image of PINN(left) and GKPINN(right), and below is the error of PINN(left) and GKPINN(right) to the exact solution.

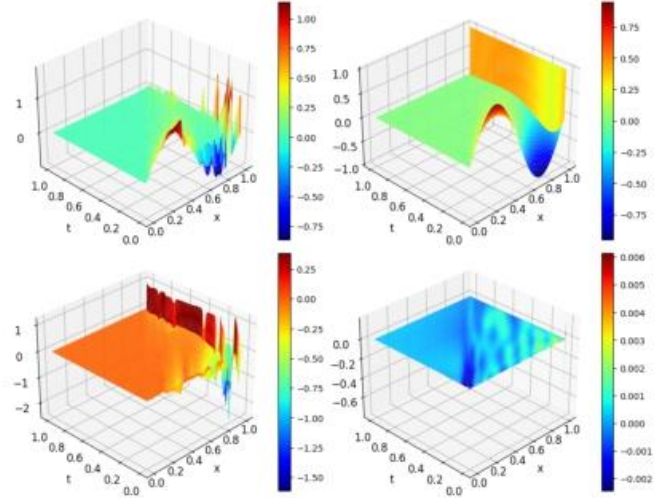

FIG. 11. When ε = 0.001 in Eq. (35), above is the prediction image of PINN(left) and GKPINN(right), and below is the error of PINN(left) and GKPINN(right) to the exact solution.

Example8

Ultimately, we proceed to examine the scenario wherein the boundary layer is positioned at the specific coordinate $x = 1$:

- $u_t - \varepsilon u_{xx} + u_x + 5u = 0, (x,t) \in (0, 1) \times (0, 1]$
- $u(x, 0) = sin(2\pi x)$
  $u(0,t) = 0, u(1,t) = 1$ (35)

Compared to Example 7, the position of the boundary layer was adjusted to the right boundary by modifying the coefficient of $u_x$. We introduced a multiplication of $e^{(x-1)/\varepsilon}$ by $u_1$ to address this issue.

The specific impact is illustrated in Figure 11, revealing

| | PINN | | GKPINN | | | |
|---|---|---|---|---|---|---|
| $\varepsilon$ | $1 \times 10^{-3}$ | | $1 \times 10^{-3}$ | | $1 \times 10^{-38}$ | |
| Experiments | Loss | L2 loss | Loss | L2 loss | Loss | L2 loss |
| Example1 | 3.319e-01 | 7.213e-01 | 2.481e-08 | 2.208e-05 | 7.242e-08 | 3.241e-05 |
| Example2 | 6.377e-01 | 4.218e-01 | 5.019e-08 | 5.704e-05 | 2.431e-08 | 2.968e-05 |
| Example3 | 6.382e-01 | 8.587e-01 | 1.272e-07 | 1.565e-05 | 9.127e-08 | 1.754e-05 |
| Example4 | 1.721e-01 | 4.992e-01 | 3.653e-07 | 1.541e-03 | 3.170e-07 | × |
| Example5 | 2.220e-01 | 4.682e-01 | 1.944e-07 | 1.466e-03 | 2.508e-07 | × |
| Example6 | 2.874e-01 | 8.412e-01 | 1.645e-06 | 4.489e-03 | 1.071e-06 | × |
| Example7 | 4.391e-01 | 4.741e-01 | 2.751e-06 | 1.947e-03 | 2.038e-06 | × |
| Example8 | 1.038e+00 | 3.422e-01 | 5.865e-06 | 7.661e-03 | 1.102e-06 | × |

TABLE I. Loss, L2 loss on the test set of all experiments.

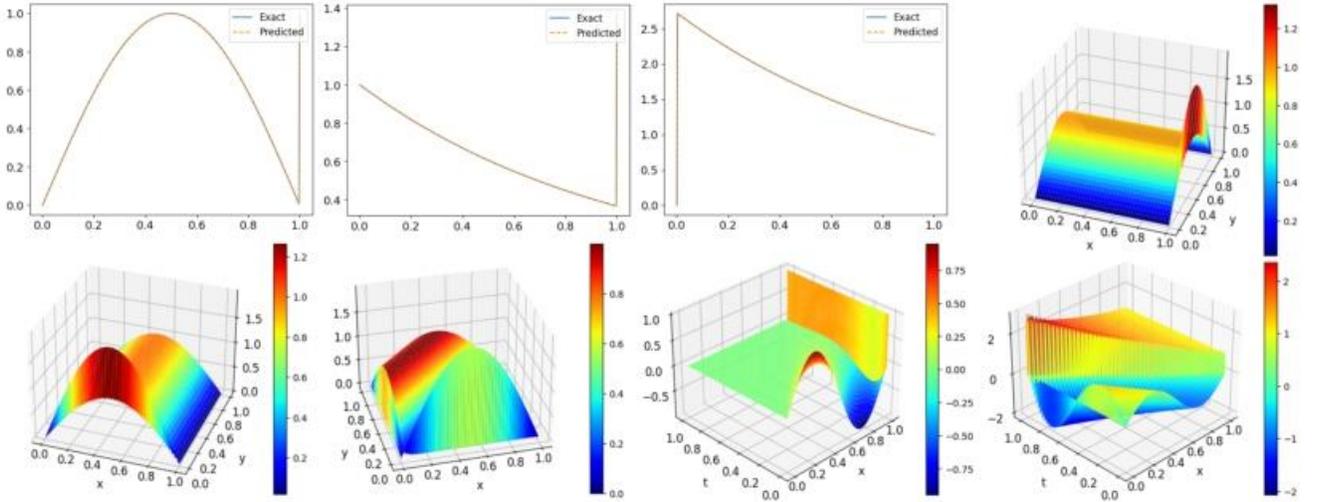

FIG. 12. When $\varepsilon = 1 \times 10^{-38}$, the results of PINN and GKPINN in Various Experiments.

a significant disparity between the two models at the $x = 1$ boundary layer.

C. Generalization

This section aims to assess the generalization capability of GKPINN when applied to SPDEs. We adjusted the perturbation parameter $\varepsilon$ to $10^{-38}$ and conducted experiments on the 8 examples mentioned earlier. The results and losses are depicted in Figure 12 and Figure 13.

As the perturbation parameter is adjusted to $10^{-38}$, the conventional method fails to accurately obtain the test set for verifying the accuracy of model training due to the absence of analytical solutions. By observing the images and Loss, it can still be demonstrated that GKPINN exhibits an excellent fitting effect on the boundary layer when dealing with SPDEs, indicating excellent generalization.

V. CONCLUSION

This study proposes the General-Kindred Physics-Informed Neural Network (GKPINN) for Singular Perturbation Differential Equations(SPDEs). GKPINN incorporates the concept of asymptotic analysis into its neural network architecture, allowing for prior knowledge about the location of the boundary layer to be obtained. By leveraging the insights from asymptotic analysis, the network is divided into multiple parts to separately handle the smooth and layer parts of SPDEs. This approach addresses a key limitation of traditional PINN, which struggles to learn steep gradients at the boundary layer. Numerical experiments demonstrate that GKPINN achieves strong convergence and accuracy, particularly in one-dimensional scenarios. Furthermore, experiments involving extreme cases of perturbation parameter $\varepsilon$ confirm GKPINN's exceptional generalization capabilities when faced

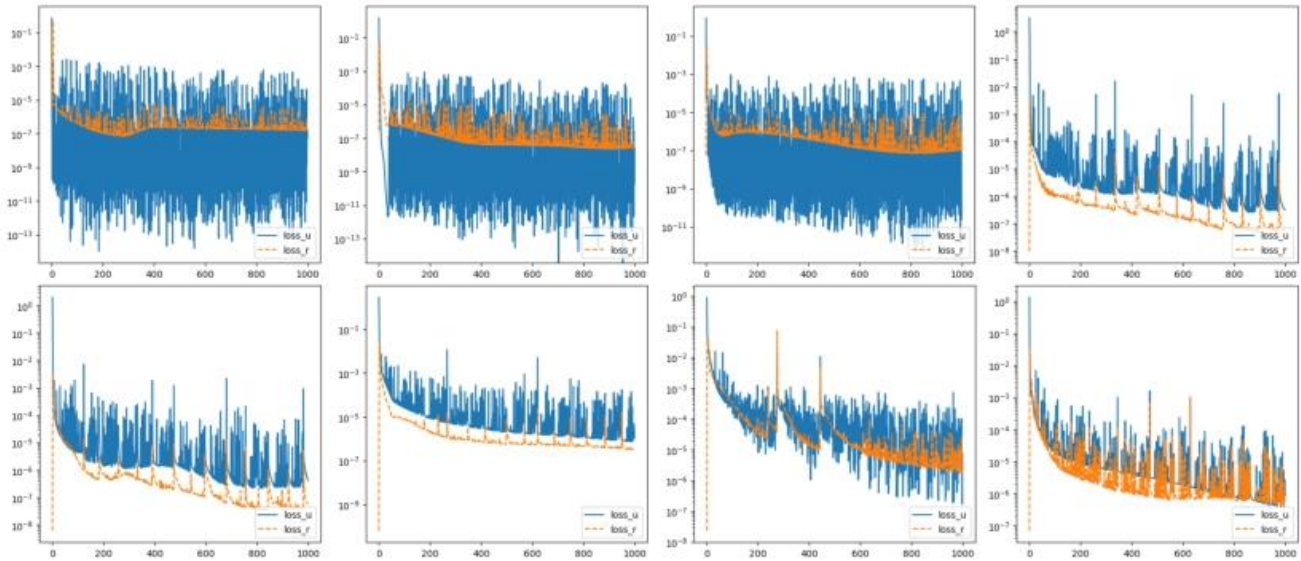

FIG. 13. When $\varepsilon = 1 \times 10^{-38}$, the losses of PINN and GKPINN in Various Experiments.

with SPDEs.